\documentclass{INTERSPEECH2023}

\usepackage{subfigure}
\usepackage{balance}
\usepackage{multirow}

\interspeechcameraready

\title{ContextSpeech: Expressive and Efficient Text-to-Speech \\ for Paragraph Reading}
\name{Yujia Xiao$^1$, Shaofei Zhang$^2$, Xi Wang$^2$, Xu Tan$^2$, Lei He$^2$, Sheng Zhao$^2$, Frank K. Soong$^2$, Tan Lee$^1$}
\address{
  $^1$Department of Electronic Engineering, The Chinese University of Hong Kong, Hong Kong\\
  $^2$Microsoft, China}
\email{yujiaxiao@link.cuhk.edu.hk, \{shazh, xwang, xu.tan, helei, Sheng.Zhao, frankkps\}@microsoft.com, tanlee@ee.cuhk.edu.hk}

\newcommand{\eg}{\textit{e}.\textit{g}.}
\newcommand{\ie}{\textit{i}.\textit{e}.} 
\newcommand{\model}{ContextSpeech}

\begin{document}

\maketitle

\begin{abstract}

\noindent While state-of-the-art Text-to-Speech systems can generate natural speech of very high quality at sentence level, they still meet great challenges in speech generation for paragraph / long-form reading. Such deficiencies are due to i) ignorance of cross-sentence contextual information, and ii) high computation and memory cost for long-form synthesis. To address these issues, this work develops a lightweight yet effective TTS system, ContextSpeech. Specifically, we first design a memory-cached recurrence mechanism to incorporate global text and speech context into sentence encoding. Then we construct hierarchically-structured textual semantics to broaden the scope for global context enhancement. Additionally, we integrate linearized self-attention to improve model efficiency. Experiments show that ContextSpeech significantly improves the voice quality and prosody expressiveness in paragraph reading with competitive model efficiency. Audio samples are available at: https://contextspeech.github.io/demo/ \\\vspace{-0.10in}

\end{abstract}
\noindent\textbf{Index Terms}: Text-to-Speech, Contextual Modeling

\section{Introduction}
\label{sec:intro}

Deep learning is powerful for speech representation learning and has shown great results on Text-to-speech (TTS) tasks~\cite{oord2018parallel,kalchbrenner2018efficient}. Representative neural network-based acoustic models in TTS evolve from autoregressive structures (\eg, Tacotron~\cite{wang2017tacotron,shen2018natural}, Deepvoice~\cite{ping2017deep}, TransformerTTS~\cite{li2019neural}) to non-autoregressive frameworks (\eg, FastSpeech~\cite{ren2019fastspeech,ren2020fastspeech}, GlowTTS~\cite{kim2020glow}) to achieve high quality generation efficiently. \nocite{xiao2023contextspeech}Recent end-to-end TTS models~\cite{kim2022vits,tan2022natural} develop the framework converting text to waveform directly without relying on an external vocoder~\cite{oord2016wavenet,kumar2019melgan,kong2020hifi}. Despite their effectiveness, we argue that existing manner of sentence-level speech synthesis is still insufficient to provide high-quality paragraph reading, in which the synthesized audio is created in paragraph-level, like news reading, audiobook, audio content dubbing, or even dialogue composed by multiple interrelated sentences.

The key reason is that most TTS models fail to capture global context among sentences within the paragraph in synthesizing audio. They usually convert text to speech in sentence-level and concatenate them for paragraph reading. An underlying fact is omitted that: sentences within the paragraph are not isolated and have various dependencies with respect to speech and textual context. Regarding the large context variation in long-form content, concatenating synthesized speech sentence by sentence has noticeable performance gap to natural recording in paragraph reading from perceptual evaluation. Additionally, the imbalanced distribution of TTS corpus data with variable-length sentences, making it difficult for TTS systems to generate high quality synthesized speech for exceptionally long or short sentences. Leaving this fact untouched, previous modeling of sentence-level context for speech synthesis has key limitations:

\begin{itemize}

\item Correlation between adjacent sentences. For paragraph reading, adjacent sentences influence each other naturally as the semantic information flowing. Thus, sentence-level speech synthesis lacks context coherence within the paragraph, and can hardly provide expressive paragraph reading.

\item Efficiency or consistency on extra-long sentences. Synthesizing extra-long sentences usually
leads to unstable results (e.g. bad alignment between text and speech) and high latency.
Generally, such sentences are partitioned into segments and then synthesized
separately, which may cause inconsistent speech rate or prosody.

\item Quality on extra-short sentences. With the data scarcity of extra-short sentences (\eg, consisted by one or two words) in corpus, TTS easily sacrifices the performance on such pattern with bad pronunciation or extremely slow speech rate.

\end{itemize}

In light of the above limitations, this work aims to study the paragraph TTS by exploring the global-level semantic dependency across different sentences. By doing so, the information transfer is enabled among sentences with variable lengths. Having realized the vital role of global context-enhanced paragraph TTS, it may suffer from scalability issue when performing speech synthesis on long paragraphs with complex cross-sentence dependency modeling. To tackle the challenges, we propose \model\ and make the following contributions:

\begin{figure*}[t!]
    \centering
    \includegraphics[width=0.95\textwidth]{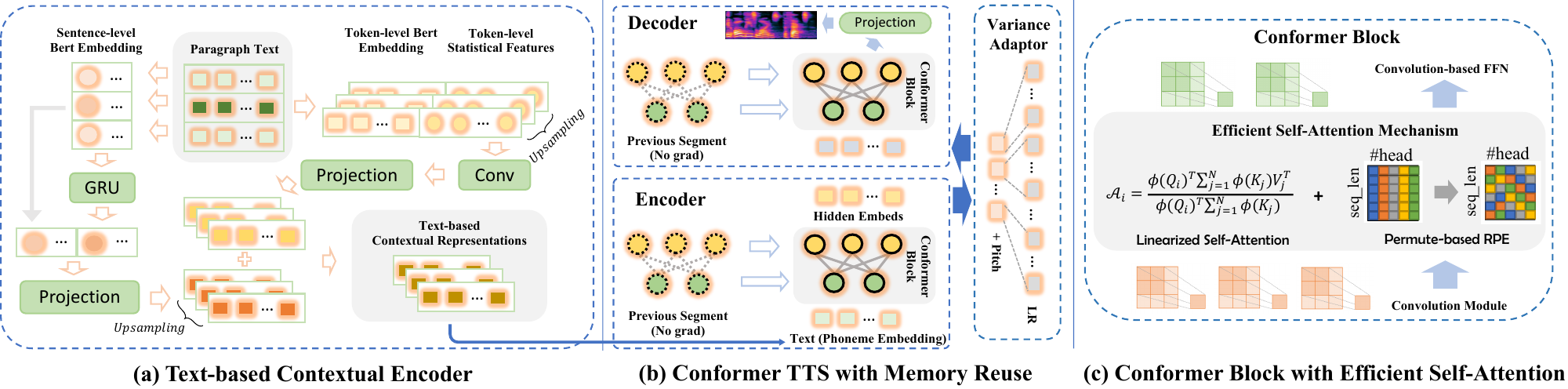}
    \vspace{-0.05in}
    \caption{The overall model architecture of our ContextSpeech with key modules}
    \label{fig:framework}
    \vspace{-0.15in}
\end{figure*}

\begin{itemize}

\item To preserve cross-sentence dependency from model perspective, a memory-cached recurrence mechanism is incorporated to transfer knowledge between segments based on the cached hidden state. We use one of the state-of-the-art sentence-level speech synthesis architecture, Conformer~\cite{peng2021conformer} based TTS in~\cite{liu2021delightfultts}, as our backbone model. The cached hidden state of each Conformer block in both encoder and decoder brings text and speech information from the previous segment.

\item Inspired by the context-aware conversational TTS~\cite{guo2021conversational}, we propose a new text-based contextual encoder to broaden the model horizon from sentence to paragraph. In particular, the proposed contextual encoder takes text-based features (\eg, BERT~\cite{devlin2018bert}-based embedding, pre-defined statistical textual information) as input and integrate them with phoneme embedding. Such integration covers information from history to future and alleviate the one-to-many mapping issue in TTS.

\item To reduce the memory and computation cost, we integrate the linearized self-attention with permute-based relative position encoding under our memory reused framework, so as to avoid quadratic complexity caused by softmax self-attention.

\end{itemize}

\noindent Experiments are carried out on a speech corpus of Chinese audiobook. The results show that ContextSpeech can generate more expressive and coherent paragraph audios compared with baseline ConformerTTS model in terms of objective and subjective evaluation. From the observation, it also alleviates the issues caused by extra-long and extra-short sentences obviously. Additionally, the final model largely alleviate the efficiency issue of extra-long input compared with baseline model.

\section{Methodology}
\label{sec:solution}

In this section, we present the details of our \model\ model whose architecture overview is shown in Figure~\ref{fig:framework}. \vspace{-0.1in}

\subsection{ConformerTTS with Memory Reuse}

\subsubsection{\bf Backbone Model} Our TTS framework is built upon the backbone model ConformerTTS, which adopts the Conformer Block (CB) in both encoder and decoder~\cite{liu2021delightfultts} of a FastSpeech2-like framework. As shown in Figure~\ref{fig:framework}-(c), the CB integrates a Convolution Module (ConvM) and a Multi-Head Self-Attention (MHSA) to model the local correlation and the global interaction. Additionally, a Convolution based Feed-Forward Network (ConvFFN) is attached after the self-attention for encoding the correlation between adjacent hidden states. More precisely, the ConvM is composed of four stacked components, including a convolutional feed-forward module, a gated linear unit (GLU), a depthwise convolution module and another convolutional feed-forward module. Let N be the number of CB stacked in encoder (or decoder), the input feature of the n-th CB is represented as $H_t^n = [h_{t,1},...,h_{t,L}]$, where $t$ is the index of current sequence and $L$ is the sequence length. In summary, the overall framework of baseline model used in this paper 1) are demonstrated in Figure~\ref{fig:framework}-(b) by ignoring the cached information 2) and consumes a softmax-based MHSA~\cite{vaswani2017attention} in CB . \vspace{-0.05in}

\subsubsection{\bf Segment-level Memory Reuse}

Inspired by ~\cite{dai2019transformer}, we cache the hidden state of previous segment in each layer and reuse it with current segment for involving contextual information, as shown in Figure~\ref{fig:framework}-(b). Notice that, the preceding segment is configured with a fixed length while a complete sentence is used as the current segment. By doing so, we can retain more intact semantic and acoustic information from both text and speech. Instead of reusing the input feature of MHSA, we choose to cache the input feature of CB directly since the ConvM can help in capturing the contextual information around the concatenation point. As the output of the $n$-th block is the input of the ($n+1$)-th block when $n < N$, the hidden state can be represented as Eq.(1), where $SG(\cdot)$ means stop-gradient and the notation [$A\circ B$] indicates concatenating hidden sequences $A$ and $B$ along the length dimension. 
\begin{equation}
H_t^{n+1} = [SG(H_{t-1}^{n+1}) \circ ConformerBlock (H_t^{n})]
\end{equation}

\subsection{Text-based Contextual Encoder}
Given the same sentence with different context, prosody of the generated speech would be different. Modelling contextual information by incorporating external linguistic and semantic features would benefit the TTS voice quality ~\cite{guo2021conversational,lei2022towards,xu2021improving,xue2022paratts}. In this section, we introduce a text-based contextual encoder to enhance the prosody expressiveness and coherence for paragraph reading. The framework is illustrated in Figure~\ref{fig:framework}-(a). Given a paragraph with a predefined context range $c$ (sentence number in a paragraph), the contextual encoder processes it to extract two kinds of contextual representations as described below:

\begin{itemize}

\item \textbf{Token-based contextual representation}. The current sentence is used to extract token- level\footnote{In our token extraction, for Chinese, "token" means "character", for English, “token” means “subword”.} Bert~\cite{devlin2018bert} embedding (TBE) and token-level statistical features (TSF). The token-level statistical features are listed in Table~\ref{tab:Token_stat_feat}, where $k$, $s$ and $p$ denote token, sentence and paragraph. For example, $i_{k\_s}$ means the index of current token in the sentence, $n_{s\_p}$ means the number of sentence in the original paragraph text, and $max(n_{k,s})$ means the maximum token number in a sentence over the training data. After concatenation, the TBE and TSF will be up-sampled and go through convolution and projection layers to align with phoneme-level features. 

\item \textbf{Sentence-based contextual representation}. For each sentence in the input paragraph, the sentence-level Bert embedding is extracted to construct a paragraph-level contextual representation (PCR) by GRU. After that, the concatenation of PCR and the current sentence embedding is fed into a projection layer and then up-sampled to phoneme-level.

\end{itemize}

\noindent The generated token-based and sentence-based contextual embedding will be added into the phoneme embedding of current sentence. With the above design, our contextual encoder not only broadens the horizon of current phoneme to global paragraph context by incorporating paragraph-level statistical features, but also improves the encoder expressiveness with phoneme embedding enhanced hierarchical contextual features. \vspace{-0.05in}

\begin{table}[h]
\vspace{-0.1in}
\caption{Token-level statistical features.}
\vspace{-0.1in}
\centering
\scriptsize
\setlength{\tabcolsep}{1.5mm}
\begin{tabular}{|c|c|c|c|c|c|}
\hline
F0 & F1 & F2 & F3 & F4 & F5 \\
\hline
$\frac{ i_{k,s} }{ n_{k,s} }$ & $\frac{ i_{k,p} }{ n_{k,p} }$ & $\frac{ i_{s,p} }{ n_{s,p} }$ & $\frac{ n_{k,s} }{ max(n_{k,s}) }$ & $\frac{ n_{k,p} }{ max(n_{k,p}) }$ & $\frac{ n_{s,p} }{ max(n_{s,p}) }$ \\
\hline
\end{tabular}
\label{tab:Token_stat_feat}
\vspace{-0.1in}
\end{table}

\subsection{Efficient Self-Attention Mechanism}
The self-attention module brings the effectiveness but also limits the model efficiency due to the quadratic time and memory complexity. Efficient Transformers~\cite{child2019generating, kitaev2020reformer, wang2020linformer, katharopoulos2020transformers} are proposed to improve the model efficiency on long-form input. Linearized self-attention is a kernel based method that can significantly reduce the computation time and memory footprint.
\subsubsection{\bf Linearized Self-Attention}
Let $X\in \mathbb{R}^{L \times d}$ be the input of self-attention module, $Q=W_q \cdot X$, $K=W_k \cdot X$ and $V=W_v \cdot X$ are linear transformations on the $X$. The canonical softmax-based self-attention mechanism can be presented as $\mathcal{A}(Q,K,V) = softmax(QK^T/\sqrt{d})V$, where the time and memory complexity is quadratic according to the input length.
\noindent Refer to~\cite{katharopoulos2020transformers}, the attention matrix can be generalized as a similarity function of $Q_i$ and $K_j$, the $i$-th or $j$-th row of the matrix $Q$ and $K$, as Eq.(2). The similarity function can be any other attention functions that are non-negative.
\begin{equation}
\mathcal{A}(Q_i,K,V) = \frac{ \sum_{j=1}^L sim(Q_i,K_j) V_j }{ \sum_{j=1}^L sim(Q_i, K_j) }
\end{equation}
 \noindent Given a qualified kernel function $\phi(x)$, the generalized row-wise attention matrix can be rewritten as Eq.(3). According to the associative property of matrix multiplication, $\phi(Q_i)^T$ can be taken out of the summation formula both in numerator and denominator as Eq.(4). Thus, we can compute the summation formula part in advance and reuse them for each query.
\begin{equation}
\mathcal{A}(Q_i,K,V)  = \frac{\sum_{j=1}^L \phi(Q_i)^T \phi(K_j) V_j}{\sum_{j=1}^L \phi(Q_i)^T \phi(K_j)}
\end{equation}
\begin{equation}
 = \Big ( \phi(Q_i)^T \sum_{j=1}^L \phi(K_j) V_j \Big ) / \Big ( \phi(Q_i)^T \sum_{j=1}^L \phi(K_j) \Big )
\end{equation}

\subsubsection{\bf Permute-based Relative Position Encoding}
To endow the linearized self-attention with the awareness of relative positional information, we applied a permute-based relative position encoding as in~\cite{chen2021permuteformer}. Particularly, the $sim(Q_i, K_j)$ in Eq.(2) will be converted to permute based format as Eq.(5). $r$ is set as 1 to avoid exploding as the sequence length increases. A premutation $B$: \{1,2,...,$d$\} $\rightarrow$ \{1,2,...,$d$\} is generated randomly, where $d$ is the dimension of query or key. Here, the first \{1,2,...,d\} and the second \{1,2,...,d\} can be treated as index collections with different order. $P_B$ is the premutation matrix of $B$, where $P_{B, ij}=1$ if $B(i)=j$; otherwise $P_{B, ij}=0$.
\begin{equation}
sim_p(Q_i, K_j) = \Big ( r_i P_B^i \phi(Q_i) \Big )^T \Big ( r^{-j} P_B^j \phi(K_j) \Big )
\end{equation}

\section{Evaluation}
\label{sec:eval}

\subsection{\bf Experimental Setup}
\vspace{-0.05in}
\noindent \textbf{Dataset}. We perform experiments on an expressive Chinese male voice. The dataset is an audiobook corpus composed of around 70 hours ($ \sim $35, 000 sentences) of narration speech and the corresponding text transcripts. We left out 100 paragraphs from the same book for objective evaluation and construct 3 different paragraph test sets from other books for subjective evaluation. Set-A: 50 paragraphs with sentences in normal length, which are used to evaluate the overall model performance on paragraph reading. Set-B: 50 paragraphs with sentences of extra-short length, i.e., one or two words, to see if the model alleviate the robustness issue in extra-short sentences. Set-C: 10 paragraphs with incremental sentence number from 2 to 11, to test the model efficiency on extra-long input sentences. \\\vspace{-0.12in}

\noindent \textbf{Model Configuration}. The ConformerTTS related model configuration is consistent to the settings in~\cite{liu2021delightfultts}. The cached memory length is set as 128 and 64 for encoder and decoder, respectively. For the text-based contextual encoder, the context size $c$ is set as 11, \ie, 5 sentences before and after the current sentence. The [input\_dim, output\_dim, kernel\_size] of the convolution layer is [774, 384, 5], which is followed by RELU, layer norm, dropout with rate 0.5 and a transformation layer with dimension $\mathbb{R}^{384 \times 384}$. The GRU layer with dimension $\mathbb{R}^{384 \times 384}$ is followed by a RELU, dropout with rate 0.5 and a linear layer with size $\mathbb{R}^{768 \times 384}$. The kernel function used in linearized self-attention is $\phi(x)=elu(x)+1$. We used MelGAN as the vocoder to generate audio from mel-spectrograms.\\\vspace{-0.12in}

\noindent \textbf{Evaluation Protocol}. We conduct paragraph MOS (mean opinion score) test to evaluate the overall voice performance of our method considering both recording and baseline model. 25 native speakers listen to each audio and give a score in 10-point scale on the overall performance and 8 specific metrics. Paragraph CMOS (comparative mean opinion score) test is used to compare the proposed model with the baseline model on different test sets. 15 native speakers listen to the synthesized samples from two models, compare them side by side and give a score from -3 to +3, where the baseline model is set as 0 for reference. Additionally, we propose a group of objective metrics to evaluate model performance according to recordings with the same transcripts, including pitch, intensity, duration and pause. For model efficiency evaluation, we conduct training on 8 NVIDIA V100 GPUs and inference on 1 NVIDIA Tesla K80 GPU.

\begin{figure}[t]
    \centering
    \includegraphics[width=0.48\textwidth]{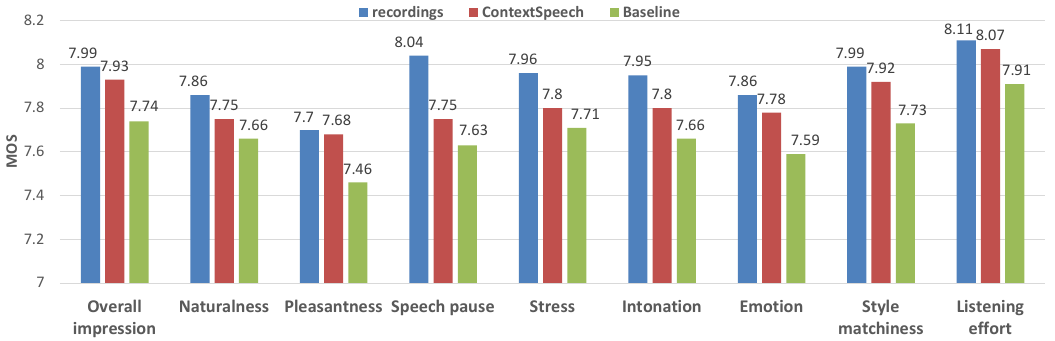}
    \vspace{-0.25in}
    \caption{Subjective Evaluation: Paragraph MOS results.}
    \label{fig:paragraph_MOS}
    \vspace{-0.1in}
\end{figure}

\begin{table}[t]
\caption{Objective Evaluation: Prosody-related metrics.}
\vspace{-0.1in}
\centering
\footnotesize
\setlength{\tabcolsep}{1.0mm}
\begin{tabular}{c|cccc}
\hline
& \multicolumn{4}{c}{Correlation}   \\ \cline{1-1} \cline{2-5}
Metrics & Pitch & Intensity & Duration & Pause \\ 
\hline
Baseline & 0.688 & 0.853 & 0.764 & 0.888 \\
ContextSpeech & 0.716 & 0.870 & 0.817 & 0.929 \\
\hline
\end{tabular}
\label{tab:Objective_metrics}
\vspace{-0.2in}
\end{table}

\subsection{\bf Quality on Paragraph Reading}
\vspace{-0.05in}
\noindent \textbf{Subjective Evaluation}. We conduct a paragraph MOS test on Set-A for our model along with baseline and recording. Figure~\ref{fig:paragraph_MOS} shows the result in terms of overall impression and other 8 specific metrics. The ContextSpeech outperforms the baseline model in all cases and achieves high-quality speech close to the recording in term of overall impression (7.93@7.99). Specifically, the proposed model reduces the MOS gap with recording from 0.25 to 0.06 compared with baseline model, which is around 76\% improvement. Especially for voice pleasantness, emotion, style matchiness and listening effort, ContextSpeech shows significant improvement with more than 50\% MOS gap reduction for expressive paragraph reading. \\\vspace{-0.12in}

\noindent \textbf{Objective Evaluation}. Besides the subjective evaluation, we also calculate the prosody-related objective metrics to measure the similarity between synthesized voice and 100 paragraph recordings. Table~\ref{tab:Objective_metrics} shows that ContextSpeech achieves improvement in each objective metric compared with baseline model, which also verifies the model performance superiority in paragraph-level prosody expressiveness.

\begin{figure}[t]
	\centering
	\subfigure[Mel-Spectrograms of Baseline sample]{
		\includegraphics[width=1.0\columnwidth]{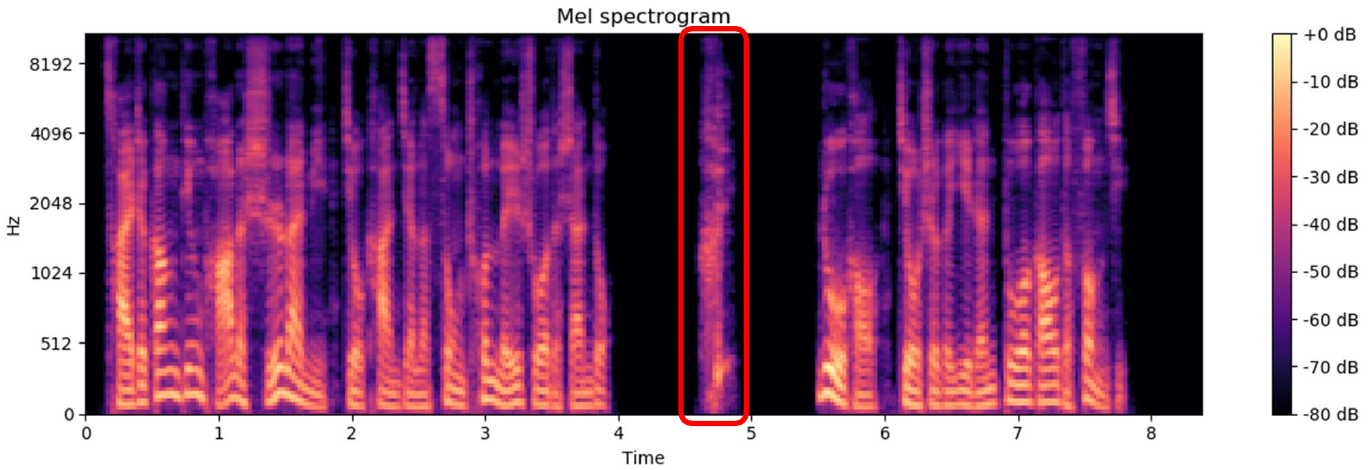}
		\label{fig:MelSpec_baseline}
	}
	\subfigure[Mel-Spectrograms of ContextSpeech sample]{
		\includegraphics[width=1.0\columnwidth]{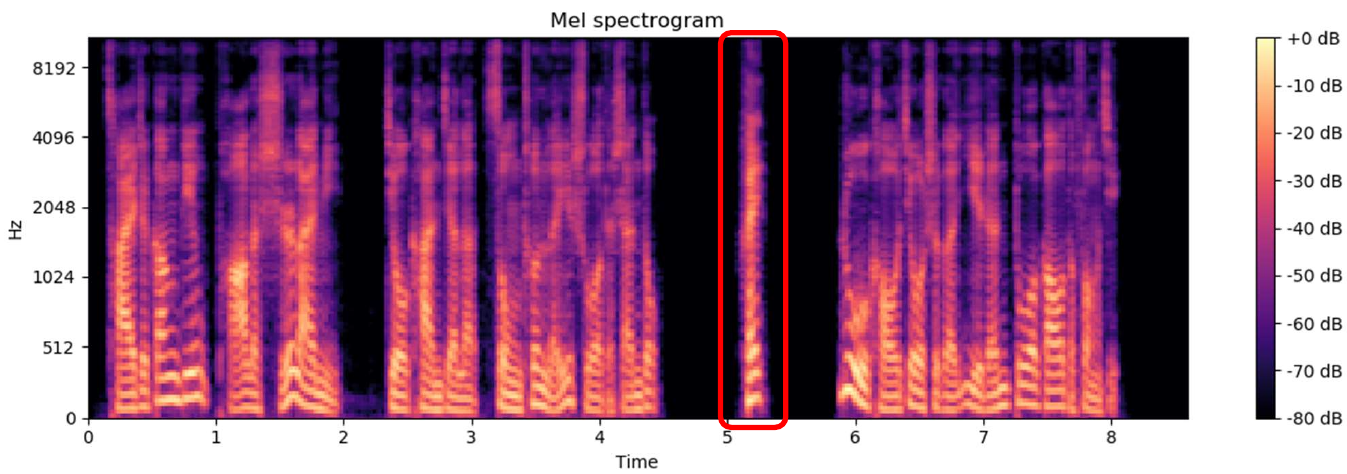}
		\label{fig:MelSpec_ContextSpeech}
	}
	\vspace{-0.1in}
	\caption{Mel-Spectrogram samples of paragraph with one-word sentence generated by \model\ and the baseline.}
	\label{fig:Mel_sepc}
        \vspace{-0.12in}
\end{figure}

\subsection{\bf Robustness on Extra-short Sentences}
 \vspace{-0.05in}
As mentioned in Section 1, extra-short sentences (one or two words) handled by sentence-level speech synthesis model usually suffer from the robustness issue, such as bad pronunciation and low speech rate. Therefore, we conduct paragraph CMOS test on Set-B. Setting the score of Baseline model as 0 for reference, \model\ obtains 0.107 CMOS gain. Both the bad pronunciation issue in one-word sentences and low speech rate issue in two-word sentences are effectively alleviated. Figure~\ref{fig:Mel_sepc} shows the mel-spectrogram samples to compare \model\ and baseline in handling one-word sentences. The red rectangles mark the position of the sentence with only one word in the paragraph. It is obvious that the spectrogram of baseline model in that position is muffle (Figure~\ref{fig:MelSpec_baseline}), while that of ContextSpeech model is much clearer with complete formant (Figure~\ref{fig:MelSpec_ContextSpeech}). By listening to the audios, we also notice that the pronunciation of that one-word sentence is distorted in baseline paragraph but clear in the \model\ paragraph.

\begin{table}[t]
\caption{Inference latency measured by millisecond per phone in different lengths of input sequences.}
\vspace{-0.1in}
\centering
\scriptsize
\setlength{\tabcolsep}{1.2mm}
\begin{tabular}{c|c|c|c|c|c|c}
\hline
\#Sent(\#Phone)& 3(414) & 5(620) & 7(898) & 9(1354) & 10(1506) & 11(1574)\\
\hline
Baseline & 0.521 & 0.422 & 0.416 & 0.609 & 0.797 & OOM\\
\hline
\multirow{2}{*}{ContextSpeech} & 0.150 & 0.111  & 0.089  & 0.083 & 0.078 & 0.075\\
 & (x3.47) & (x3.80) & (x4.67) & (x7.34) & (x10.22) & \\
\hline
\end{tabular}
\label{tab:Inference_latency}
\vspace{-0.2in}
\end{table}

\begin{table}[t]
\caption{CMOS Test on Paragraphs with Extra-Short or Long sentences for Voice Quality Comparison.}
\vspace{-0.1in}
\centering
\small
\setlength{\tabcolsep}{1.0mm}
\begin{tabular}{c|c|c}
\hline
& Baseline & ContextSpeech \\
\hline
Paras with Extra-short Sentences & 0 & +0.107\\
\hline
Paras with Extra-long Sentences & 0 & +0.226\\
\hline
\end{tabular}
\label{tab:Extra-Short-Long}
\vspace{-0.05in}
\end{table}

\subsection{\bf Efficiency on Extra-long Sentences}
\vspace{-0.05in}
The efficient self-attention module described in Section 2.3 largely improves the model efficiency. For training stage, \model\ with linearized self-attention achieves 2x of speedup and 2x of memory tolerance compared with using softmax-based self-attention. For inference stage, \model\ shows significant efficiency superiority over the baseline especially for extra-long inputs. Table~\ref{tab:Inference_latency} illustrates the inference latency for baseline and ContextSpeech model according to different input phoneme length on Set-C. The baseline model run into out-of-memory when the input phone number increase to 1574. In contrast, \model\ is able to handle such long sequences. Additionally, \model\ outperforms the baseline in each group and achieves more than 10x speedup when the input length is 1506. Furthermore, we perform paragraph CMOS on this test set and obtain 0.226 CMOS gain (Table~\ref{tab:Extra-Short-Long}). In summary, for extra-long input sentence, ContextSpeech shows better expressiveness and efficiency compared with baseline. 

\begin{table}[t]
\caption{Ablation Study with Paragraph CMOS Test.}
\vspace{-0.1in}
\centering
\small
\setlength{\tabcolsep}{1.5mm}
\begin{tabular}{c|c|c|c}
\hline
ContextSpeech & - MR & - TCE & - ESA \\
\hline
0 & -0.085 & -0.048 & -0.030 \\
\hline
\end{tabular}
\label{tab:Ablation_study}
\vspace{-0.2in}
\end{table}

\subsection{\bf Model Ablation Study}
We conduct ablation study to evaluate the effectiveness of key modules in \model. Table~\ref{tab:Ablation_study} shows the paragraph CMOS results on Set-A component-wise ablation results.\\\vspace{-0.12in}

\noindent \textbf{Memory Recurrence (MR)}. Memory reuse mechanism described in Section 2.1 is proposed to enlarge the receptive field of current segment to see more historical information. To verify its effectiveness, we remove it from \model model and do a paragraph CMOS test for comparison. Set \model model as 0 for reference, removing MR cause -0.085 regression, which demonstrates the contribution from MR mechanism. \\\vspace{-0.12in}

\noindent \textbf{Text-based Contextual Encoder (TCE)}. As described in Section 2.2, we proposed a text-based contextual encoder to leverage hierarchical contextual information from plain paragraph text. To evaluate its effectiveness, we do paragraph CMOS test to compare the models with and without TCE module. The negative score -0.048 verifies the positive effect of TCE module. \\\vspace{-0.12in}

\noindent \textbf{Efficient Self-Attention Mechanism (ESA)}. ESA is introduced in Section 2.3, which aims to improve model efficiency and robustness especially on extra-long input. The efficiency improvement and corresponding performance on extra-long input are proved in Section 3.4. Here we replace the ESA in \model\ by softmax-based self-attention with relative position encoding in Transformer-XL, to evaluate the performance in paragraphs with normal-length sentence. The paragraph CMOS result, -0.030, demonstrates that the ESA module will not cause quality regression and even with slight improvement.

\section{Conclusion}
\label{sec:conclusion}

In this paper, we propose ContextSpeech, which is an expressive and efficient TTS model for generating speech of paragraph reading. The memory reuse mechanism is introduced in the encoder-decoder framework to incorporate historical information of text and speech to current sentence. Text-based contextual information is encoded in a hierarchical structure to extend the model capability to paragraph level. Furthermore, linearized self-attention with compatible relative position encoding is adopted to improve the model efficiency. Experiments on Chinese audiobook corpus demonstrate that ContextSpeech achieved superior voice quality and expressiveness in paragraph reading compared with the baseline model, 76\% reduction on the MOS gap to recording. ContextSpeech also shows robustness performance on extra-short sentences with 0.107 CMOS gain, and improves both the expressiveness (0.226 CMOS gain) and efficiency ($ \sim $10x speedup) over the extra-long sequences.

\clearpage
\balance
\bibliographystyle{IEEEtran}
\bibliography{mybib}

\end{document}